\documentclass[conference]{IEEEtran}
\IEEEoverridecommandlockouts
\usepackage{cite}
\usepackage{amsmath,amssymb,amsfonts}
\usepackage{algorithmic}
\usepackage{graphicx}
\usepackage{textcomp}
\usepackage{xcolor}
\usepackage{booktabs}

\newcommand{\modelname}{UniUSNet}
\newcommand{\OrganNum}{7}
\newcommand{\AnnoNum}{9.7K}
\newcommand{\ImageNum}{6.9K}

\begin{document}

\title{\modelname: A Promptable Framework for Universal Ultrasound Disease Prediction and Tissue Segmentation
\thanks{\textsuperscript{*}Corresponding Author: Tao Tan. Email: taotan@mpu.edu.mo}
}

\author{\IEEEauthorblockN{1\textsuperscript{st} Zehui Lin}
\IEEEauthorblockA{\textit{Faculty of Applied Sciences} \\
\textit{Macao Polytechnic University}\\
Macao SAR, China \\
p2316858@mpu.edu.mo}
\and
\IEEEauthorblockN{2\textsuperscript{nd} Zhuoneng Zhang}
\IEEEauthorblockA{\textit{Faculty of Applied Sciences} \\
\textit{Macao Polytechnic University}\\
Macao SAR, China \\
p2316955@mpu.edu.mo}
\and
\IEEEauthorblockN{3\textsuperscript{rd} Xindi Hu}
\IEEEauthorblockA{\textit{Shenzhen RayShape Medical Technology Co. Ltd.} \\
Shenzhen, China \\
h15851832081@163.com}
\and
\IEEEauthorblockN{4\textsuperscript{th} Zhifan Gao}
\IEEEauthorblockA{\textit{School of Biomedical Engineering} \\
\textit{Sun Yat-sen University}\\
Shenzhen, China \\
gaozhifan@gmail.com}
\and
\IEEEauthorblockN{5\textsuperscript{th} Xin Yang}
\IEEEauthorblockA{\textit{School of Biomedical Engineering} \\
\textit{Shenzhen University}\\
Shenzhen, China \\
xinyang@szu.edu.cn}
\and
\IEEEauthorblockN{6\textsuperscript{th} Yue Sun}
\IEEEauthorblockA{\textit{Faculty of Applied Sciences} \\
\textit{Macao Polytechnic University}\\
Macao SAR, China \\
yuesun@mpu.edu.mo}
\and
\IEEEauthorblockN{7\textsuperscript{th} Dong Ni}
\IEEEauthorblockA{\textit{School of Biomedical Engineering} \\
\textit{Shenzhen University}\\
Shenzhen, China \\
nidong@szu.edu.cn}
\and
\IEEEauthorblockN{8\textsuperscript{th} Tao Tan\textsuperscript{*}}
\IEEEauthorblockA{\textit{Faculty of Applied Sciences} \\
\textit{Macao Polytechnic University}\\
Macao SAR, China \\
taotan@mpu.edu.mo}
}

\maketitle

\begin{abstract}
    Ultrasound is widely used in clinical practice due to its affordability, portability, and safety. However, current AI research often overlooks combined disease prediction and tissue segmentation. We propose \modelname, a universal framework for ultrasound image classification and segmentation. This model handles various ultrasound types, anatomical positions, and input formats, excelling in both segmentation and classification tasks. Trained on a comprehensive dataset with over \AnnoNum~annotations from \OrganNum~distinct anatomical positions, our model matches state-of-the-art performance and surpasses single-dataset and ablated models. Zero-shot and fine-tuning experiments show strong generalization and adaptability with minimal fine-tuning. We plan to expand our dataset and refine the prompting mechanism, with model weights and code available at (https://github.com/Zehui-Lin/UniUSNet).
\end{abstract}

\section{Introduction}

\begin{figure}[htbp]
    \centerline{\includegraphics[width=0.5\textwidth]{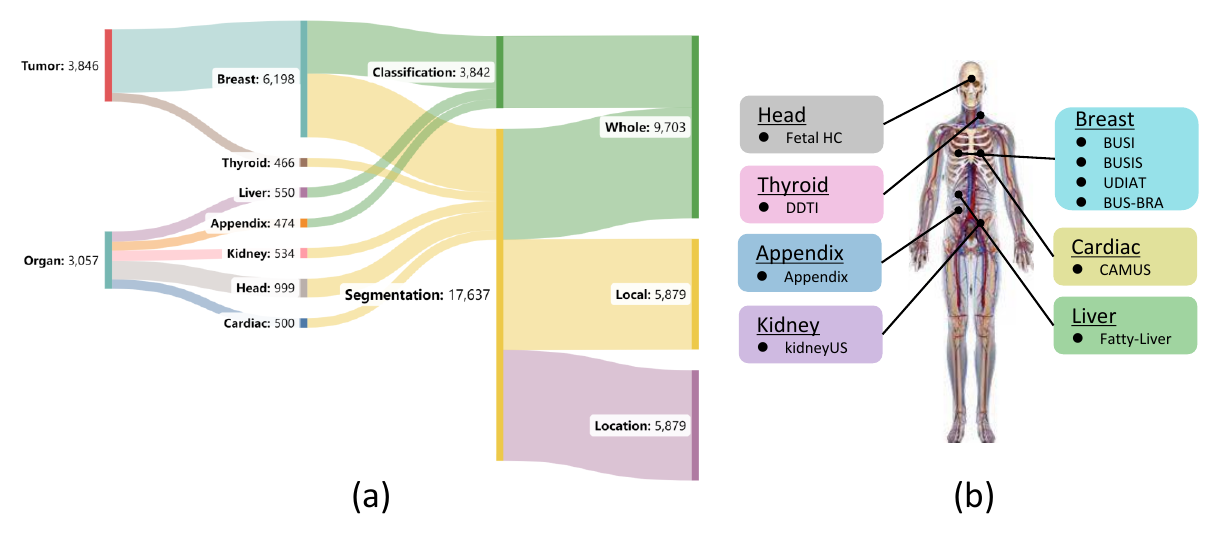}}
    \caption{The BroadUS-\AnnoNum~dataset contains \AnnoNum~annotations of \ImageNum~ultrasound images from \OrganNum~different anatomical positions. (a) The number of effective instances corresponding to nature of the image, position, task and input type. Note that a breast image can contain both segmentation and classification labels, and an image with segmentation labels can form three different input types (b) The different anatomical positions, and their corresponding public dataset abbreviations.}
    \label{fig:dataset}
\end{figure}

Ultrasound imaging is crucial in clinical settings due to its affordability, safety, and portability. While deep learning models for ultrasound have progressed, they often have narrow applications and are constrained by limited annotated data. Recently, General Medical Artificial Intelligence (GMAI) models \cite{zhang2023challenges} have shown strong generalization across diverse tasks and datasets, enhancing diagnostic accuracy and efficiency. However, a universal GMAI model specifically for ultrasound remains unexplored, highlighting the need for a comprehensive ultrasound-specific dataset that considers the unique characteristics of ultrasound images.

Recent advancements in AI models for ultrasound imaging have addressed tasks like breast tumor prediction~\cite{huang2021aw3m}, thyroid tumor segmentation~\cite{manh2022multi} and prenatal image quality control~\cite{he2021statistical}. However, these models struggle with new tasks or datasets, requiring extensive retraining and overlooking intrinsic dataset-task relationships. The focus is now on universal models, such as Segment Anything Model (SAM) \cite{kirillov2023segment}, known for its versatility in segmentation tasks across domains. Adaptations like MedSAM \cite{ma2024segment} and SAM-Med2D \cite{cheng2023sam} show promise in medical imaging. SAMUS \cite{lin2023samus} is a specialized version of SAM tailored for ultrasound.

Current methods focus on segmentation but struggle with classification, crucial in clinical settings. SAM's fixed architecture limits adding classification features, whereas our Swin-Unet-based multi-branch network integrates classification more effectively. For ultrasound's complexity, we propose fully automated prompts that encode prior knowledge, unlike existing methods like CLIP embeddings \cite{zhang2023continual} and HyperConv \cite{han2024synthesis}, which enhance performance but lack synchronous training and remain underexplored for ultrasound.

We propose \modelname, a universal, promptable framework for ultrasound that addresses multiple clinical tasks. Our contributions include: (1) Curating a comprehensive dataset, BroadUS-\AnnoNum, with over \ImageNum~images and \AnnoNum~annotations across \OrganNum~anatomical positions, ensuring robust validation. (2) Developing a versatile transformer-based model using four prompts—image nature, anatomical position, task, and input type—enhancing adaptability without extensive retraining. (3) Demonstrating \modelname's effectiveness through experiments on BroadUS-\AnnoNum, showing superior performance over single-dataset models and those lacking prompts, while also achieving comparable results to state-of-the-art models. (4) Conducting zero-shot and low-cost fine-tuning experiments, underlining the model's strong generalization and adaptability to new clinical tasks.

The contributions of this work include:
\begin{itemize}
\item Comprehensive Dataset Collection: We compiled extensive ultrasound datasets covering \OrganNum~anatomical positions with over \AnnoNum~annotations for robust model validation.
\item Versatile Model Framework: Our transformer-based framework uses four model prompts to enhance flexibility across various clinical tasks.
\item Extensive Experiments: The model outperforms single-dataset and ablated versions, achieving state-of-the-art performance and strong generalization to new domains.
\end{itemize}

\section{Methodology}
\begin{figure}[htbp]
    \centering
    \includegraphics[width=0.5\textwidth]{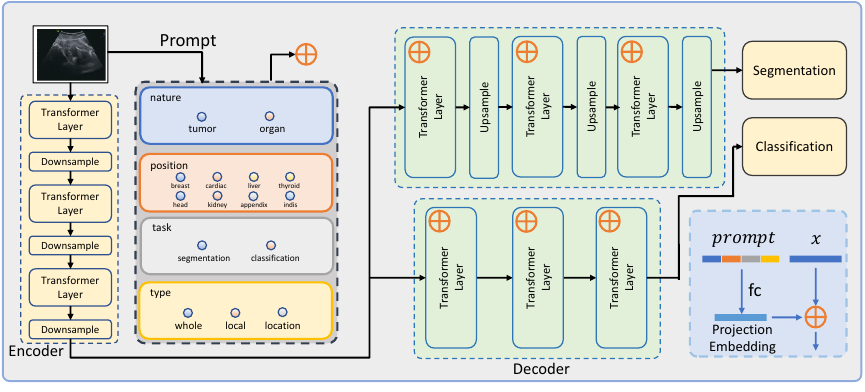}
    \caption{The architecture of the proposed \modelname.}
    \label{fig:framework}
\end{figure}

The architecture of \modelname~(Fig.~\ref{fig:framework}) is a general encoder-decoder model that uses prompts to simultaneously handle multiple ultrasound tasks like segmentation and classification. The encoder extracts features, while task-specific decoders are enhanced by four types of prompts—nature, position, task, and type—added to each transformer layer via prompt projection embedding, boosting the model's versatility and performance.

\subsection{Uni-Transformer Learning Framework}

The \modelname~framework is designed for multitasking on ultrasound images, built on a modified Swin-Unet~\cite{cao2022swin} with one encoder and two decoders for segmentation and classification. The segmentation decoder includes a skip connection from the encoder, while the classification decoder omits the upsampling layer but maintains the same depth for computational equivalence.

Inspired by \cite{zhao2023one}, we balanced data across positions due to varying dataset sizes, optimizing the model's learning of domain-specific knowledge. Using curriculum learning, we prioritized segmentation data before classification, leveraging edge and location info to enhance classification performance. For training, dice and cross-entropy losses are used for segmentation, while classification relies on cross-entropy loss.

\subsection{Prompting Content}

We use four prompts—\textbf{nature}, \textbf{position}, \textbf{task}, and \textbf{type}—to provide diverse information about images and tasks, enhancing the model's flexibility and interpretability. \textbf{Nature Prompt}: The nature prompt identifies whether the image is of an organ or a tumor. Organ prompts focus on structural regularity, while tumor prompts highlight heterogeneous information. \textbf{Position Prompt}: The position prompt indicates the image's anatomical location, helping the model apply relevant domain knowledge. Options include breast, cardiac, liver, thyroid, head, kidney, appendix, and a general "Indis" prompt for unseen organs. \textbf{Task Prompt}: The task prompt specifies the model's focus—either segmentation (delineating structures) or classification (categorizing the image). \textbf{Type Prompt}: The type prompt defines the input image type: whole (entire image), local (cropped region), or location (brightness-enhanced areas), allowing the model to adjust spatial granularity for both local and global tasks.

\subsection{Prompting Strategy}
Inspired by Vision Transformers (ViT) \cite{dosovitskiy2020image}, our prompt integration strategy is straightforward. We define four prompt one-hot vectors, concatenate them, and project them into the transformer's dimensions using a fully connected (fc) layer.

These projected prompts are added to the feature $x$ at each decoder layer, ensuring full involvement in task processing. Each layer has its own fc layer for independent adaptation. The learnable fc layer optimizes the prompts during feature learning, enhancing network reusability. For new tasks, we only need to update the prompts without adding new branches.

The implementation is as follows:
\begin{equation}
    \begin{split}
    x^{'} &= x + fc_{i,k}(concat(p_{nature}, p_{position}, p_{task}, p_{type})), \\
    &\quad i=1,2,3, k=1,2
    \end{split}
\end{equation}

where $x_{i,k}$ is the input feature, $p$ is the prompt vector, and $fc$ is the fully connected layer for layer $i$ and branch $k$.

\subsection{Adaptive Domain Testing and Fine-tuning with Adapters}

To assess our universal model's adaptability to new datasets, we introduced a fine-tuning approach inspired by adapter theory \cite{jia2022visual}. Instead of retraining the entire model, we insert small trainable modules (adapters) into the existing model layers. These adapters allow the model to adjust to new tasks without altering core parameters, preserving the original knowledge.

In our method, we freeze the encoder-decoder structure and only fine-tune the fully connected (fc) projection layers linked to the prompts. This approach minimizes computational overhead and prevents catastrophic forgetting, enabling our model to efficiently adapt to new tasks and domains while maintaining robust versatility.

\section{Experimental Results}

\subsection{Dataset and Implementation Details} 

As shown in Table~\ref{tab:data_table}, the dataset was split into training, validation, and testing sets (7:1:2), ensuring all images from the same patient were in the same partition. Segmentation loss combined 0.4 cross-entropy and 0.6 dice loss. We trained models for 200 epochs using AdamW (learning rate: 3e-4) with data augmentation techniques like random flipping, rotation (-20° to 20°), and cropping. Implementation was done in PyTorch, using an NVIDIA A4000 GPU. Segmentation was evaluated with the dice coefficient, and classification with accuracy, both ranging from 0 to 1.

\begin{table}[htbp]
    \centering
    \caption{BroadUS-\AnnoNum~Dataset description.}
    \label{tab:data_table}
    \resizebox{0.45\textwidth}{!}{%
    \begin{tabular}{@{}cccc@{}}
    \toprule
    Dataset     & Position & Image Num & Annotation                   \\ \midrule
    BUSI \cite{al2020dataset}        & Breast   & 780       & Classification, Segmentation \\
    BUSIS \cite{zhang2022busis}       & Breast   & 562       & Segmentation                 \\
    UDIAT \cite{yap2017automated}       & Breast   & 163       & Classification, Segmentation \\
    BUS-BRA \cite{gomez2023bus}     & Breast   & 1875      & Classification, Segmentation \\
    Fatty-Liver \cite{byra2018transfer} & Liver    & 550       & Classification               \\
    kidneyUS \cite{singla2023open}    & Kidney   & 534       & Segmentation                 \\
    DDTI \cite{pedraza2015open}        & Thyroid  & 466       & Segmentation                 \\
    Fetal HC \cite{van2018automated}   & Head     & 999       & Segmentation                 \\
    CAMUS \cite{leclerc2019deep}       & Cardiac  & 500       & Segmentation                 \\
    Appendix \cite{marcinkevics_regensburg_2023}    & Appendix & 474       & Classification               \\ \bottomrule
    \end{tabular}
    }
\end{table}

\subsection{Comparison with SOTA and Ablation Study}

\begin{table}[htbp]
    \centering
    \caption{Overall performance comparison on BoardUS-\AnnoNum~dataset.}
    \label{tab:exp_table1}

    \resizebox{0.45\textwidth}{!}{

    \begin{tabular}{@{}ccccccc@{}}
    \toprule
    Dataset             & Task        & \begin{tabular}[c]{@{}c@{}}SAM\\ (Interactive)\end{tabular}     & \begin{tabular}[c]{@{}c@{}}SAMUS\\ (Interactive)\end{tabular}   & \begin{tabular}[c]{@{}c@{}}Single\\ (Automatic)\end{tabular}  & \begin{tabular}[c]{@{}c@{}}\modelname~w/o prompt\\ (Automatic)\end{tabular}  & \begin{tabular}[c]{@{}c@{}}\modelname~\\ (Automatic)\end{tabular}  \\ \midrule
    \multicolumn{2}{c}{Params}        & 90.49M  & 130.10M          & \begin{tabular}[c]{@{}c@{}}29.59M / 34.98M\\(347.07M)\end{tabular} & 86.26M & 86.29M \\ \midrule
    BUS-BRA             & seg         & 17.89\% & \textbf{83.88\%} & 80.44\% & 73.99\% & 75.59\% \\
    BUSIS               & seg         & 28.63\% & \textbf{88.99\%} & 86.62\% & 87.39\% & 87.84\% \\
    CAMUS               & seg         & 53.96\% & 72.48\% & 89.80\% & 91.16\% & \textbf{92.05\%} \\
    DDTI                & seg         & 23.65\% & 69.70\% & \textbf{74.13\%} & 61.15\% & 66.70\% \\
    Fetal\_HC           & seg         & 60.27\% & \textbf{97.60\%} & 85.36\% & 95.97\% & 96.69\% \\
    kidneyUS            & seg         & 35.24\% & 67.54\% & 80.51\% & \textbf{82.29\%} & 80.23\% \\
    UDIAT               & seg         & 40.17\% & \textbf{84.35\%} & 75.02\% & 59.91\% & 61.00\% \\ \midrule
    \multicolumn{2}{c}{Seg Average}   & 37.12\% & 80.65\% & \textbf{81.70\%} & 78.84\% & 80.01\% \\ \midrule
    Appendix            & cls         & /       & /       & \textbf{66.32\%} & 53.68\% & 52.84\% \\
    BUS-BRA             & cls         & /       & /       & 73.07\% & 78.67\% & \textbf{84.69\%} \\
    Fatty-Liver         & cls         & /       & /       & 81.82\% & 90.91\% & \textbf{92.36\%} \\
    UDIAT               & cls         & /       & /       & 69.70\% & 87.88\% & \textbf{88.79\%} \\ \midrule
    \multicolumn{2}{c}{Cls Average}   & /       & /       & 72.72\% & 77.78\% & \textbf{79.67\%} \\ \midrule
    \multicolumn{2}{c}{Total Average} & /       & /       & 78.43\% & 78.46\% & \textbf{79.89\%} \\ \bottomrule
    \end{tabular}

    }
\end{table}

Given our model's general-purpose design, we compared it with state-of-the-art (SOTA) models, particularly SAM and its variants. We evaluated our model against SAM-based methods in two configurations: using SAM's official pre-trained weights and a SAM variant for ultrasound data, SAMUS, trained on our BroadUS-\AnnoNum~Dataset. Additionally, we performed ablation studies. We first trained individual models for each dataset (Single), highlighting the traditional approach's lack of generalizability. Then, we trained our \modelname~model without prompts to assess whether prompts genuinely enhance performance.

Table \ref{tab:exp_table1} shows that SAM's official weights perform poorly in zero-shot inference (37.12\%) due to the domain gap between natural and medical images. SAMUS improves performance (80.65\%) but doesn't surpass the Single model, likely due to dataset heterogeneity. Our automatic prompt model, with 66\% fewer parameters, achieves similar segmentation results (80.01\%). Ablation studies reveal that \modelname~(79.89\%) outperforms both the ablation version (78.46\%) and Single (78.43\%) models, proving the effectiveness of prompts. While \modelname~and \modelname~w/o prompt models have fewer parameters, they excel in classification over segmentation, possibly due to the network's multi-branch structure, suggesting a need for more balanced learning.

\begin{figure}[htbp]
    \centering
    \includegraphics[width=0.4\textwidth]{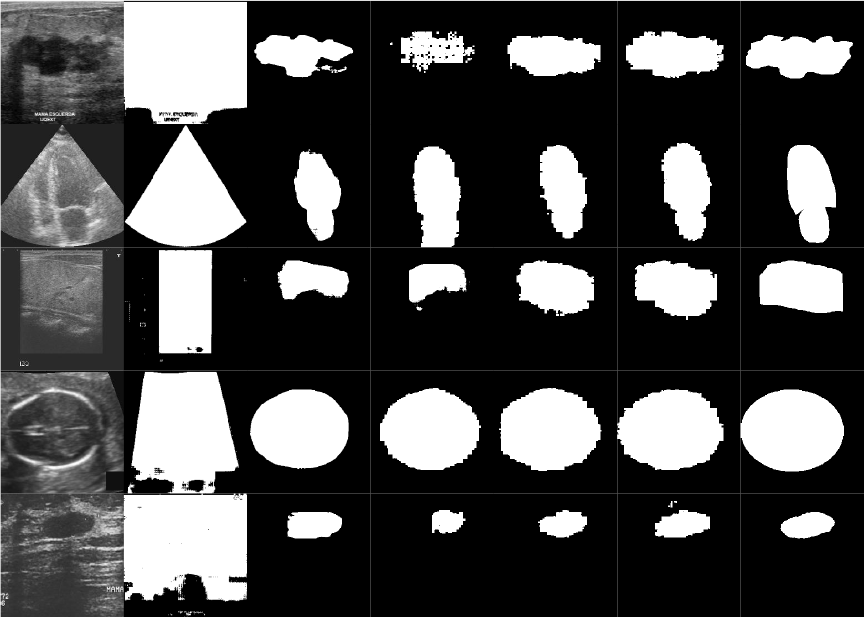}
    \caption{Some examples of segmentation result. Each column From left to right: original image, SAM, SAMUS, Single, \modelname~w/o prompt, Prompt and ground truth.}
    \label{fig:seg}
\end{figure}

\begin{figure}[htbp]
    \centering
    \includegraphics[width=0.4\textwidth]{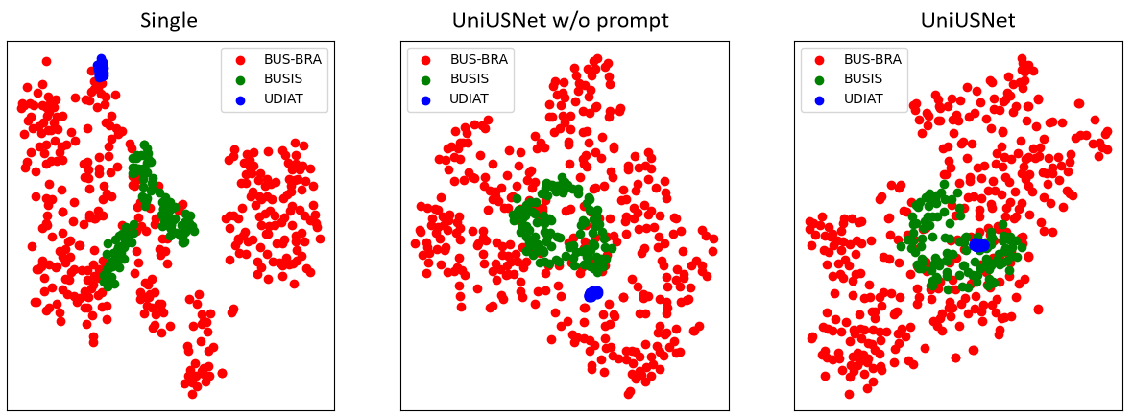}
    \caption{t-SNE visualization.}
    \label{fig:tsne}
\end{figure}

Using t-SNE \cite{van2008visualizing}, we visualized feature distributions of the BUS-BRA, BUSIS, and UDIAT datasets. Figure \ref{fig:tsne} shows that the Single model has a clear domain shift, while \modelname~w/o prompt reduces this shift, indicating better domain adaptation. Prompts further minimize the domain offset, aligning with Table \ref{tab:exp_table2}. Segmentation results in Figure \ref{fig:seg} reveal that \modelname~outperforms SAM and other models by effectively using nature and position prompts for deeper task understanding.

\subsection{Evaluating Adapter Performance on New Datasets}

\begin{table}[htbp]
    \centering
    \caption{Adapter performance comparison on BUSI dataset.}
    \label{tab:exp_table2}
    \resizebox{0.45\textwidth}{!}{

    \begin{tabular}{ccccccc}
    \hline
    Dataset & Task & Single  & \begin{tabular}[c]{@{}c@{}}\modelname\\w/o prompt\end{tabular}  & \modelname & Scratch  & Adapter             \\ \hline
    BUSI    & seg  & 60.85\% & 66.90\% & 66.91\% & 68.89\% & \textbf{69.56\%} \\
    BUSI    & cls  & 80.80\% & 77.60\% & 78.40\% & 85.90\% & \textbf{86.40\%} \\ \hline
    \multicolumn{2}{c}{Average}  & 70.83\% & 72.25\% & 72.66\% & 77.40\% & \textbf{77.98\%} \\ \hline
    \end{tabular}%
    
    }
\end{table}

We tested our model's generalization on the BUSI dataset by excluding it from training and performing zero-shot inference using the Single, \modelname~w/o prompt, and \modelname~models. The results (Table \ref{tab:exp_table2}) show that \modelname~w/o prompt and \modelname~outperform the Single model, demonstrating better generalization and prompt effectiveness. Additionally, the Adapter setup, with minimal fine-tuning, surpasses the Scratch setup, showcasing our model's adaptability to new datasets efficiently.

\section{Conclusion}

We propose a universal model for medical ultrasound imaging that integrates four prompts, enhancing learning from multiple perspectives. Experiments on the BroadUS-\AnnoNum~dataset validate its effectiveness and potential for fine-tuning on new datasets. Future work will expand our analysis to further assess the model's adaptability.

\section*{Acknowledgment}

This work was supported by Science and Technology Development Fund of Macao (0021/2022/AGJ) and Science and Technology Development Fund of Macao (0041/2023/RIB2).

\bibliographystyle{IEEEtran}
\bibliography{ref.bib}

\end{document}